\documentclass{article}

\usepackage{arxiv}

\usepackage[utf8]{inputenc} 
\usepackage[T1]{fontenc}    
\usepackage{hyperref}       
\usepackage{url}            
\usepackage{booktabs}       
\usepackage{amsfonts}       
\usepackage{nicefrac}       
\usepackage{microtype}      
\usepackage{lipsum}		
\usepackage{graphicx}
\usepackage{natbib}
\usepackage{doi}
\usepackage{amsmath}
\usepackage{subcaption}

\title{An LLM-Explainable DRL Framework for Passenger-
Directed Autonomous Driving}

\date{}	

\author{
\href{https://orcid.org/0009-0001-0904-8292}{\includegraphics[scale=0.06]{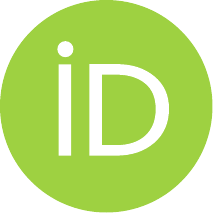}}\hspace{1mm}Ouided Braoui$^{1,2}$\\
\texttt{ouided.braoui@univ-eiffel.fr}
\And
\href{https://orcid.org/0009-0003-0886-1251}{\includegraphics[scale=0.06]{orcid.pdf}}\hspace{1mm}Meriem Bouali$^{1}$\\
\texttt{bouali@estin.dz}
\And
\href{https://orcid.org/0000-0002-0309-8942}{\includegraphics[scale=0.06]{orcid.pdf}}\hspace{1mm}Nadir Farhi$^{2}$\\
\texttt{nadir.farhi@univ-eiffel.fr}
\\[2mm]
$^{1}$ LITAN, ESTIN, RN 75, Amizour, 06300, Bejaia, Algeria
\\
$^{2}$ Cosys-Grettia, Univ Gustave Eiffel, F-77454 Marne-la-Vallée, France
}

\renewcommand{\headeright}{}
\renewcommand{\undertitle}{}
\renewcommand{\shorttitle}{An LLM-Explainable DRL Framework for Passenger-Directed Autonomous Driving}

\hypersetup{
pdftitle={An LLM-Explainable DRL Framework for Passenger-
Directed Autonomous Driving},
pdfsubject={q-bio.NC, q-bio.QM},
pdfauthor={David S.~Hippocampus, Elias D.~Striatum},
pdfkeywords={First keyword, Second keyword, More},
}

\begin{document}
\maketitle

\begin{abstract}
Autonomous vehicles offer the potential for safer and more efficient mobility, yet public trust remains limited due to the lack of transparency in their decision-making. This work addresses this issue by combining deep reinforcement learning (DRL) for adaptive driving control with large language model (LLM)-based explainability modules designed to communicate agent behavior to passengers. DRL agents were trained in simulation using a Dueling Double Deep Q-Network to follow distinct driving requests: \textit{fast}, \textit{comfort}, and \textit{stop}. They demonstrated stable learning, safe compliance with traffic rules, and reliable switching between modes within a single trip. In parallel, LLM modules were introduced to interpret passenger requests, determine when explanations were needed, and generate concise, safety-oriented justifications. Results show that this framework, serving as a proof of concept for integrating RL decision-making and LLMs, balances safety, adaptability, and explainability, and is most effective when requests are delayed or overridden due to safety constraints.
\end{abstract}

\keywords{Autonomous Vehicles, Deep Reinforcement Learning, Large Language Models, Explainability.}

\section{Introduction}

In recent years, autonomous vehicles (AVs) have evolved from experimental prototypes into a vital element of modern transportation systems. Advances in artificial intelligence have supported significant progress in perception, planning, and control tasks. Yet, despite these technological improvements, their public adoption remains constrained by societal skepticism. Recent surveys reveal that most people are still afraid of riding in a self-driving car \citep{aaa2025fear}. A major contributor to this discomfort is the lack of transparency and explainability in AV decision-making systems. As a result, passengers often have limited interaction with the vehicle and cannot understand the reasoning behind its actions, especially in critical situations.

To bridge this gap between AV performance and public trust, explainable artificial intelligence (XAI) has been proposed as a potential solution. According to \cite{yang2025explaining}, designing an interactive human–machine interface (HMI) that provides clear and comprehensible explanations of AV decision-making, covering both benevolence and competence, can enhance user trust. However, most existing XAI approaches remain developer-oriented rather than passenger-oriented, and to the best of our knowledge, no current framework links passenger voice commands with control policies and corresponding real-time explanations.

\subsection{Related Work}
 
DRL methods have shown strong performance in autonomous driving tasks including lane keeping \citep{deep_rl_freamework_for_av} and changing \citep{Bouali2025}, highway merging \citep{Wang_2017}, safe overtaking \citep{overtaking_manouvres}, and merging in occluded intersections \citep{navigating_occluded}. These approaches demonstrated high adaptability to dynamic traffic. However, neural networks deployed in DRL models typically operate as black boxes \citep{BARREDOARRIETA202082}, providing no human-understandable reasoning behind their decisions. In addition, they do not integrate real-time passenger input into the control loop, limiting their ability to adapt to user driving preferences.

Existing work on explainability in autonomous driving can be grouped into three main directions: visual model representations, safety- and rule-guided approaches, and recent LLM/VLM-based reasoning methods. In visual explainability, PolarPoint-BEV \citep{PolarPointBEV} provided interpretable bird’s-eye-view representations using a sparse polar mesh, but its limited class set and fixed radial intervals restricted descriptive richness and flexibility. Similarly, \cite{Guti_rrez_Zaballa_2024} improved segmentation explainability using hyperspectral imagery and U-Net activations on HSI-Drive, though reliance on costly hyperspectral sensors limited practical deployment. Concept-bottleneck methods \citep{bottleneck} enabled interpretable internal reasoning but depended on predefined concepts and lacked interactivity. 

Safety wise, \cite{adversarial} studied robustness to adversarial sensor attacks with saliency-based explanations, yet only within a roundabout scenario. Cog-MP \citep{cogmp} introduced a hierarchical RL framework improving transparency through cognitive, decisional, and operational layers, but its explanations remained technical and developer-oriented. Similarly, \cite{ballardini2025humanlikesemanticnavigationautonomous} used LLMs to translate passenger instructions into logical ASP rules for semantic navigation, producing traceable plans but relying on manually defined rules and high-quality instructions. 

More recently, LLM and VLM approaches have sought to improve interpretability in autonomous driving. VLAAD \citep{vlaad} combined Video Q-former with LLaMA-2 and introduced a large visual instruction dataset, but relied on non-driving-specific encoders and frontal RGB-only input. DiLu \citep{wen2024dilu} enhanced decision-making through reasoning, reflection, and episodic memory, though it suffered from latency and hallucinations. Graph-of-thought reasoning with GPT-4 \citep{chi2024multimodal} generated multimodal explanations but remained limited to monocular sensors and small-scale evaluation, while RAG-Driver \citep{yuan2024ragdriver} enabled zero-shot explanations via retrieval-augmented reasoning, constrained by model size and context limits. Closed-loop systems such as SimLingo \citep{renz2025simlingovisiononlyclosedloopautonomous} and X-Driver \citep{liu2025xdriverexplainableautonomousdriving} integrated multimodal perception with language reasoning but were validated only in simulation and faced real-time latency challenges. A comprehensive multimodal framework \citep{zarghani2025multimodalframeworkexplainableautonomous} further fused video, sensor data, and text for accurate predictions and explanations, though it remained computationally heavy and lacked real-world validation. Overall, although LLM/VLM approaches have improved interpretability, they remain limited by latency, real-time interaction, and integration with driving policies, especially RL-based systems that adapt to personalized passenger commands while providing context-aware explanations.

\subsection{Contributions}

In this work, we propose a hybrid framework that combines DRL and LLMs to create an autonomous driving system that is safe, adaptive and explainable to passengers in a simulation environment. More specifically, we build a novel approach that consists of:

\begin{enumerate}
    \item a passenger-oriented XAI pipeline integrating LLMs with DRL-based driving control,
    \item a Dueling Double Deep Q-Network (D3QN) agent capable of operating in three driving modes (fast, comfort, stop) within a simulated urban environment, and
    \item an LLM-based conflict-detection mechanism that triggers explanations when safety rules override passenger commands.
\end{enumerate}

The remainder of this paper is organized as follows: Section \ref{sec:Section2} presents the framework, Section \ref{sec:Section3} describes the RL formulation, Section \ref{sec:Section4} details the LLM design, Section \ref{sec:Section5} reports the results, and Section \ref{sec:Section6} concludes the paper.

\section{General Problem Formulation and System Overview}
\label{sec:Section2}

Since existing explainability methods mainly target developers or rely on offline video datasets, there is a need for AVs that can follow real-time passenger commands, support natural interaction, and provide clear explanations when safety constraints prevent immediate compliance. An effective AV system must therefore listen to passenger requests in real time, interpret their intent (whether explicit or implicit), map it to high-level driving commands, and verify whether the request can be safely executed under traffic and safety constraints. When a request cannot be fulfilled, the system should generate a clear and reassuring explanation for non-expert passengers. In this paper, our objective is to develop an interactive, explainable AV framework that integrates real-time passenger command understanding with reinforcement learning-based driving decisions and context-aware explanations.

\subsection{Simulation Environment and Driving Commands}

\begin{figure}[htbp]
    \centering

    \begin{subfigure}[b]{0.48\textwidth}
        \centering
        \includegraphics[width=0.6\textwidth]{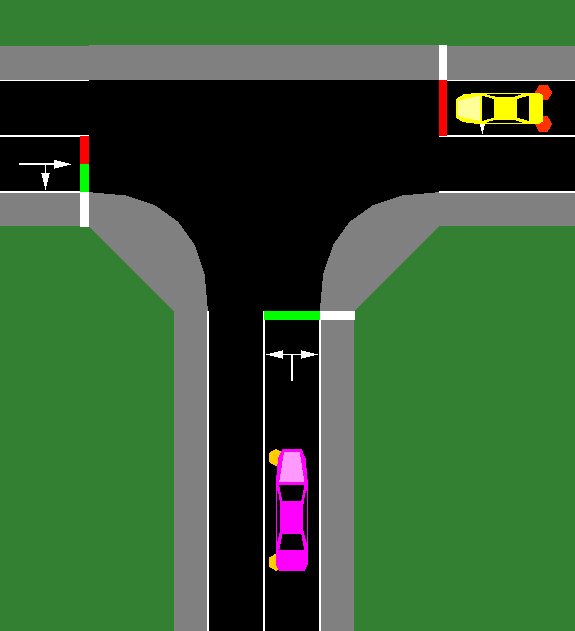}
        \caption{First signalized intersection for traffic-light compliance.}
        \label{fig:first_signalized_intersection}
    \end{subfigure}
    \hfill
    \begin{subfigure}[b]{0.48\textwidth}
        \centering
        \includegraphics[width=0.5\textwidth]{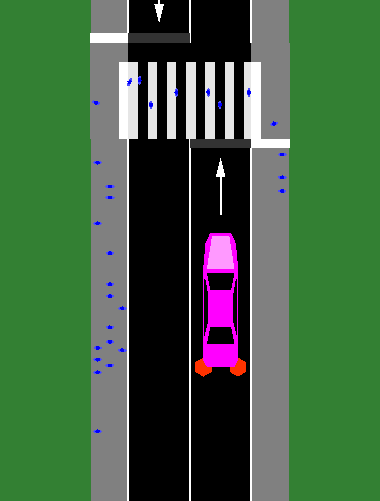}
        \caption{Unsignalized pedestrian crossing for yielding behavior.}
        \label{fig:unsignalized_pedestrian_crossing}
    \end{subfigure}

    \vspace{0.3cm}

    \begin{subfigure}[b]{0.98\textwidth}
        \centering
        \includegraphics[width=\textwidth]{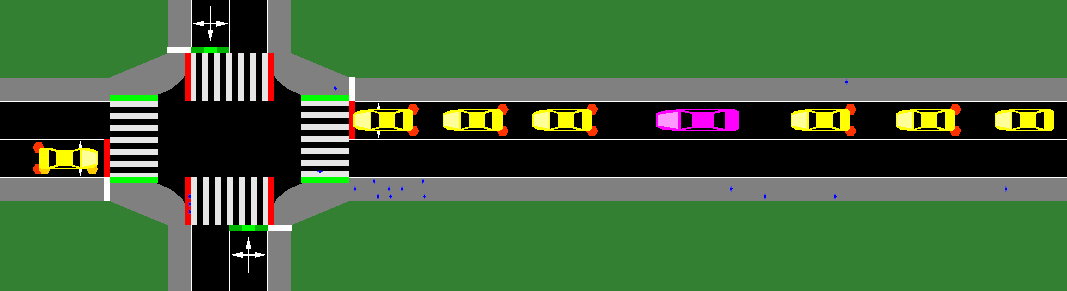}
        \caption{Second signalized intersection with leading vehicles and pedestrian crossing.}
        \label{fig:second_signalized_intersection}
    \end{subfigure}

    \caption{Urban SUMO scenario main components. The scenario includes signalized intersections, an unsignalized pedestrian crossing, leading vehicles, and pedestrian interactions.}
    \label{fig:urban_sumo_scenario}
\end{figure}

We consider a two-way urban road with one lane per direction, simulated in SUMO (Simulation of Urban Mobility) \citep{sumo} microscopic traffic simulator, representative of typical city traffic. This configuration allows us to focus on longitudinal control in this initial study, while lateral control and more complex multi-lane scenarios are left for future work. The scenario includes two signalized intersections, one unsignalized pedestrian crossing, varying speed limits, and moderate car flow. The first signalized intersection (Fig. \ref{fig:first_signalized_intersection}) and the pedestrian crossing (Fig. \ref{fig:unsignalized_pedestrian_crossing}) allow the ego vehicle to learn signal compliance and yielding behavior in isolation, while the second intersection (Fig.\ref{fig:second_signalized_intersection}) introduces leading vehicles and a coordinated pedestrian crossing, creating realistic multi-constraint situations. All background vehicles employ SUMO’s Krauss \citep{krauss} default car-following model.

Passenger requests are abstracted into three commands: \textit{fast}, \textit{comfort}, and \textit{stop}. The \textit{fast} command prioritizes progress toward the speed limit when safe, The \textit{comfort} command favors smooth and conservative driving, and the \textit{stop} command requests an immediate halt overriding other objectives.

\subsection{System Architecture and Workflow}

\begin{figure}[h]
    \centering
    \includegraphics[width=\textwidth]{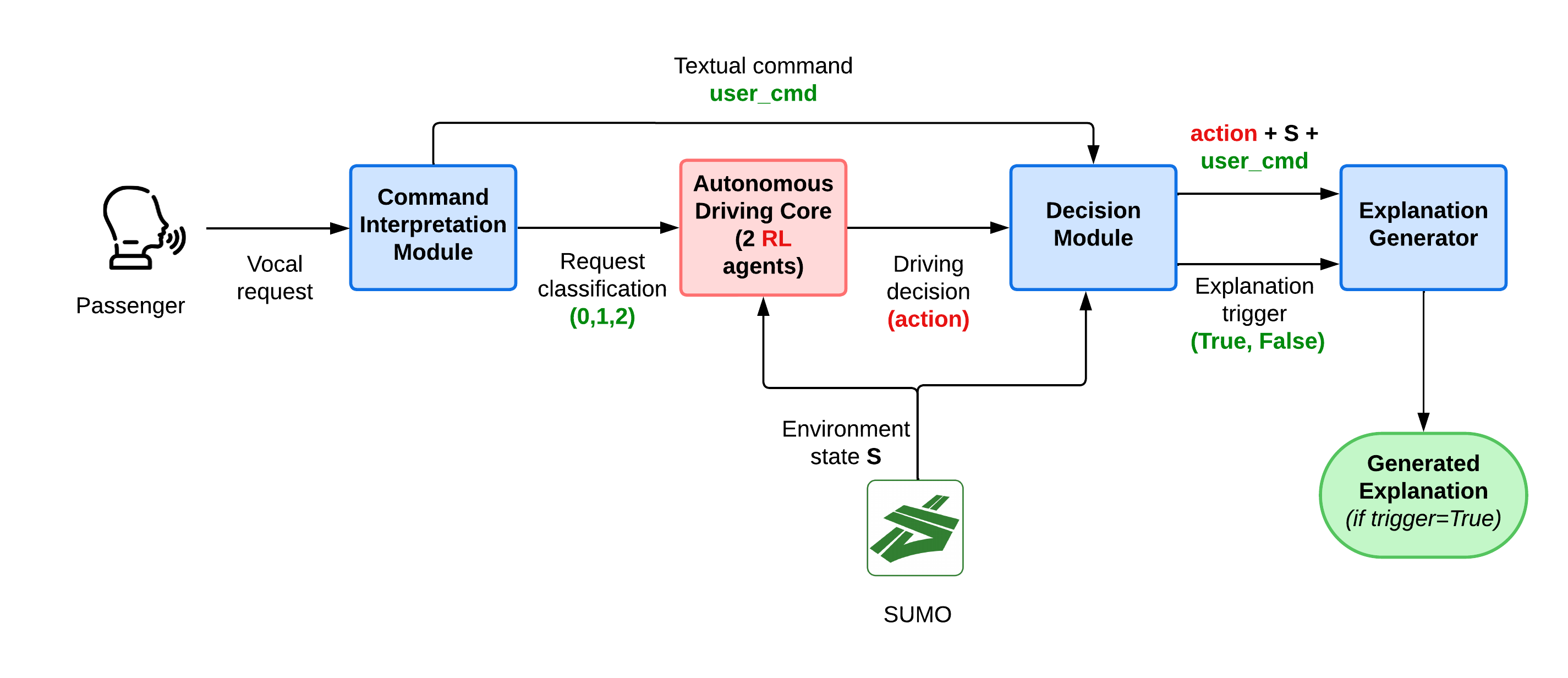}
    \caption{Block diagram of the system architecture.}
    \label{fig:architecture}
\end{figure}

The proposed architecture consists of four integrated subsystems, illustrated in Fig.~\ref{fig:architecture}, enabling real-time passenger interaction, adaptive control, and explainable autonomous driving.

The workflow begins with spoken passenger commands, classified as direct or indirect following \citet{cui2024personalizedautonomousdrivinglarge}'s taxonomy, which are captured and transcribed by the \textbf{Command Interpretation Module (LLM-1)} using Whisper \citep{whisper}, and mapped to one of three driving intents: \textit{fast}, \textit{comfort}, or \textit{stop}. This intent is passed to the \textbf{Autonomous Driving Core}, which contains two DRL agents trained for fast and comfort driving, respectively, while both obey the stop command. The active policy is selected according to the passenger’s request.

Meanwhile, SUMO provides the environment state, including ego speed, speed limit, headway to the leading vehicle, relative speed, and the distance and state of traffic lights and pedestrian crossings. The \textbf{Decision Module (LLM-2)} evaluates whether the executed action conflicts with the passenger’s request based on the current state, interpreted intent, and selected action. When a conflict is detected, the \textbf{Explainability Generator (LLM-3)} produces a clear explanation to justify the vehicle’s behavior.

All LLM modules use GPT-4.1-mini \citep{openai_gpt41mini}, enabling low-latency reasoning and real-time explanations. This architecture integrates passenger intent interpretation, DRL-based decision-making, and LLM-based explanation to provide adaptive and transparent autonomous driving.

\section{Reinforcement Learning Formulation}
\label{sec:Section3}
The autonomous driving task is modeled as a Markov Decision Process (MDP), where the agent observes the traffic state and selects discrete control actions to maximize cumulative reward. The state describes the ego vehicle and its surroundings, while the action corresponds to speed adjustment commands. The reward function encodes safety, efficiency, comfort, and traffic rule compliance. 

Due to the complexity of interactive urban traffic dynamics, we employ a D3QN to train the driving policy \citep{Ducrocq2023} . The dueling architecture improves training stability, reduces overestimation bias, and enables efficient learning from structured state features provided by the SUMO environment. The learned policy governs vehicle speed control under the three driving requests (\textit{fast}, \textit{comfort}, \textit{stop}), while ensuring compliance with traffic rules such as red lights, safe following distances, and pedestrian yielding.

\subsection{State Representation}\label{AA}

The choice of state variables is critical in our work, as it directly influences control performance and determines the information available for interpreting and explaining the agent’s behavior.
Following the car-following state formulation proposed by \citet{Zhu_2020}, we retain the core variables describing the ego vehicle and its interaction with the leading vehicle, namely the ego speed, headway distance, relative speed as well as the road speed limit. Building on this baseline, we extend the state space to include additional dimensions required for command-driven urban driving,
including the passenger command, traffic light distance and
phase, and pedestrian crossing proximity. Accordingly, the agent receives an 8-dimensional state vector that encodes both the passenger's request and the traffic context, defined as:
\begin{equation}
s = (cmd,\; v_{\text{ego}},\; v_{\text{lim}},\; d_{\text{lead}},\; \Delta v,\; d_{\text{tl}},\; tls,\; d_{\text{cross}})
\end{equation}
where $cmd$ is the passenger command label (0 = fast, 1 = comfort, 2 = stop), $v_{\text{ego}}$ is the ego-vehicle speed, $v_{\text{lim}}$ is the road speed limit, $d_{\text{lead}}$ is the headway distance to the leading vehicle, $\Delta v$ is the relative speed with respect to the leader, $d_{\text{tl}}$ is the distance to the next traffic light, $tls$ is its current state (0 = green, 1 = yellow, 2 = red), and $d_{\text{cross}}$ is the distance to the nearest detected pedestrian crossing the street.

\subsection{Action Space}
Inspired by \citep{bilateral}, which considers a continuous action space, we adopt a discretized action representation to define a compact and interpretable set of longitudinal control commands. Accordingly, our action space is discrete and consists of seven longitudinal velocity adjustment commands:
\begin{equation}
\label{eq:action_space}
A = \{-3,\,-2,\,-1,\,0,\,+1,\,+2,\,+3\}\;
\end{equation}
where each action represents a discrete change in the vehicle's speed (m/s) applied at each decision step. No lateral maneuvers are allowed, as there is only one lane for each direction. The ego vehicle is controlled by an RL agent instead of the SUMO default car-following model.

\subsection{Key Features for Reward Function}
We present the key features used to train the RL agent so that it ensures safety while aligning with the passenger’s requested driving style. These metrics, based on \citep{Zhu_2020}, are incorporated into the reward design to reflect safety, efficiency, and comfort.

\textbf{i. Time to Collision (TTC):}  
Safety is the highest priority. TTC measures the remaining time before a collision between vehicle $n$ and its leader $n-1$:
\begin{equation}
TTC(t) = -\frac{D_{n-1,n}(t)}{\Delta V_{n-1,n}(t)}
\end{equation}
where $D_{n-1,n}(t)$ is the gap distance and $\Delta V_{n-1,n}(t)$ the relative speed. To ensure collision avoidance and maintaining a safe distance from the leading vehicle, we adopt a 2-second TTC safety limit, consistent with values commonly reported in the literature (1.5–5\,s). The following shaping function penalizes low TTC values \citep{Zhu_2020}:
\begin{equation}
\label{eq:ttc}
f_{\text{TTC}}(ttc)=
\begin{cases}
\log\!\left(\frac{ttc}{2}\right), & 0 \le ttc \le 2, \\
0, & \text{otherwise}.
\end{cases}
\end{equation}
It encourages the agent to avoid small TTC values, as the logarithmic function diverges to negative infinity when \textit{ttc} approaches zero, yielding strong penalties near collision. For larger TTC values, the penalty diminishes and the function approaches zero. Incorporating this term into the reward encourages the agent to keep TTC above the 2-second threshold, improving collision avoidance and overall safety.

\textbf{ii. Time Headway:}  
Time headway measures how long it would take for the follower to reach the leader’s position:
\begin{equation}
H_n(t) = \frac{D_{n-1,n}(t) + L_{n-1}}{V_n(t)}
\end{equation}
where $D_{n-1,n}(t)$ is the vehicle gap, $V_{n-1}(t)$ the ego speed, and $L_{n-1}$ is the leader's length. This metric captures the trade-off between traffic efficiency and safety, as shorter values increase roadway capacity while excessively small headways may be unsafe. Following \citep{Zhu_2020}, time headway is modeled using a log-normal distribution fitted to empirical NGSIM car-following data, which exhibits a right-skewed shape with a peak around typical human driving values and a long tail corresponding to more conservative spacing. The probability density function is given by:

\begin{equation}
\label{eq:headway}
\begin{array}{l}
H \sim \text{LogNormal}(\mu,\sigma), \\[6pt]
f_{\text{headway}}(h) =
\frac{1}{h \sigma \sqrt{2\pi}}
\exp\!\left(
-\frac{(\ln h - \mu)^2}{2\sigma^2}
\right)
\end{array}
\end{equation}

The parameters $\mu$ and $\sigma$ were calibrated using 1,341 car-following events, yielding $\mu = 0.4226$ and $\sigma = 0.4365$. This distribution assigns the highest likelihood to moderate headway values (around $1$--$2$\,s), while both dangerously small and excessively large headways are less probable, encouraging a balance between safety and traffic efficiency.

\textbf{iii. Jerk:}  
Passenger comfort is strongly influenced by driving smoothness, as abrupt changes in acceleration are perceived as uncomfortable. Jerk captures this effect by penalizing rapid variations in acceleration:
\begin{equation}
Jerk(t) = \frac{A(t) - A(t-1)}{\Delta t}
\end{equation}
where $A(t)$ and $A(t-1)$ denote accelerations at times $t$ and $t-1$, and $\Delta t$ is the control step equal to 1s.

These three metrics: \textbf{TTC} (for safety), \textbf{headway} (for efficiency), and \textbf{jerk} (for comfort), form the core features guiding the reward function. They ensure that learned driving policies remain aligned with the passenger’s commands (fast, careful driving). 

\subsection{Reward Functions}

The reward function shapes the behavior of the driving policies. In addition to respecting traffic signals, pedestrian crossings, and ensuring safe car-following, three passenger requests are considered: fast driving, comfort-oriented driving, and emergency stop. The general reward is:
\begin{equation}
R =
\begin{cases}
R_0, & \text{if fast driving requested},\\
R_1, & \text{if comfortable driving requested},\\
R_2, & \text{otherwise (STOP command)}.
\end{cases}
\end{equation}

\subsubsection{Fast Driving Reward Design}

For fast driving, efficiency and progress are prioritized while maintaining safety. Comfort is less critical than speed and travel time. Accordingly, the reward combines adherence to the request, traffic-rule compliance, speed-limit respect, and collision avoidance:
\begin{equation}
R_0 = R_{\text{efficient\_car\_following}}
+ R_{\text{traffic\_rules}}
+ R_{\text{speed}}
+ R_{\text{collision}} .
\end{equation}

By combining formulas (\ref{eq:ttc}) and (\ref{eq:headway}), we obtain this expression encouraging efficient car-following \ref{eq:reward_efficient_car_flw}. The headway probability density is directly used as a reward-shaping term, so that headway values close to those empirically observed in NGSIM human car-following data receive higher rewards, while very small or excessively large headways are penalized:
\begin{equation}
R_{\text{efficient\_car\_following}}
= 2\cdot f_{TTC}(\text{ego\_ttc})
+ 3\cdot f_{headway}(\text{ego\_h}) .
\label{eq:reward_efficient_car_flw}
\end{equation}
The weights were selected to favor efficient car-following behavior while explicitly accounting for safety and collision avoidance.

For traffic-rule compliance, violations are strongly penalized, correct stops (e.g., at red lights or pedestrian crossings) receive a moderate reward, and unnecessary stops incur an equivalent penalty to avoid overly conservative behavior. These values were determined through empirical fine-tuning, as they provided the best balance between safety and traffic efficiency. Thus, the reward is defined as:
\begin{equation}
\label{eq:traffic_reward}
R_{\text{traffic\_rules}} =
\begin{cases}
-r_{\text{unsafe}}, & \text{if violates red light or pedestrian crossing or stops without reason}, \\
r_{\text{respect}}, & \text{if respects red light or pedestrian crossing by stopping}.\\
\end{cases}
\end{equation}

Initially, the design of $R_{\text{traffic\_rules}}$ relied on hard-coded distance thresholds that penalized the agent for failing to brake or stop near red lights or pedestrian crossings. However, this approach either caused the agent to stop too early or, when the thresholds were reduced, allowed it to bypass the penalty entirely and learn unsafe, rule-violating behaviors. These limitations indicated that discrete distance-based shaping was insufficient for consistent traffic-rule compliance, motivating the introduction of a virtual leader \citep{treiber2013} to enforce more reliable stopping behavior. In this approach, a zero-length virtual vehicle is spawned at the stop line whenever a signal turns red or yellow or a pedestrian is detected. This allows the ego vehicle to treat the situation as a car-following task, approaching the virtual leader, adjusting speed naturally, and stopping precisely at a safe distance, thereby replacing rigid distance thresholds with a more realistic and rule-consistent stopping mechanism.

In addition, the agent must respect the speed limit defined for the lane, which is enforced through the following speed-limit term:
\begin{equation}
R_{\text{speed}} =
\begin{cases}
-r_{\text{unsafe}}, & v_t > v_{\text{lim}},\\
3 \cdot \frac{v_t}{v_{\text{lim}}}, & \text{otherwise}.
\end{cases}
\end{equation}
These reward values were selected to strongly discourage speed-limit violations while encouraging efficient progress toward the permitted speed when operating within legal bounds.

A large negative penalty is assigned to collisions to reflect their critical severity and to ensure that collision avoidance dominates all other driving objectives during learning. Thus, the fast-driving agent follows strict safety constraints while encouraging efficient progress:
\begin{equation}
\label{eq:collision}
R_{\text{collision}} = -r_{\text{critical}}.
\end{equation}

\subsubsection{Comfort-Oriented Reward Design}

For comfort, the emphasis shifts to smoothness and pleasant driving. Speed maintenance is less important, and jerk is explicitly penalized. The reward is:
\begin{equation}
R_1 =
R_{\text{comfort\_car\_following}}
+ R_{\text{traffic\_rules}}
+ R_{\text{speed}}
+ R_{\text{collision}} .
\end{equation}

The weighting scheme of this driving mode reflects a comfort-oriented objective by reducing the emphasis on efficiency-related terms while penalizing high jerk to promote smoother and more gradual vehicle motion:
\begin{equation}
\begin{aligned}
R_{\text{comfort\_car\_following}}
&= 2\cdot f_{TTC}(\text{ego\_ttc})  + 1.5\cdot f_{headway}(\text{ego\_h})
- {jerk^2}.
\end{aligned}
\end{equation}

The reward for traffic rules remain unchanged as in Eq. (\ref{eq:traffic_reward}). In contrast to the fast driving mode, the speed term for this command reflects the reduced priority of speed:
\begin{equation}
R_{\text{speed}} =
\begin{cases}
-r_{\text{unsafe}}, & v_t > v_{\text{lim}},\\
1.5\cdot \frac{v_t}{v_{\text{lim}}}, & \text{otherwise}.
\end{cases}
\end{equation}

Collisions are penalized similarly to Eq. (\ref{eq:collision}).

\subsubsection{Stop Command Reward Design}

The stop command is intended for emergencies or unsafe conditions. The reward design enforces its emergency nature by strongly penalizing non-stopping actions while encouraging immediate or progressive braking behavior:
\begin{equation}
R_2 =
\begin{cases}
r_{\text{respect}}, & v_t = 0 \;(\text{stopping}),\\
-r_{\text{critical}}, & a_t \in \{0,1,2,3\}\;(\text{maintaining or increasing speed}),\\
0, & \text{otherwise (} v_t > 0 \text{ and } a_t \in \{-3,-2,-1\}\text{)}.
\end{cases}
\end{equation}

After empirical fine-tuning, three reward values were selected: $r_{\text{respect}} = 2$ for positive behavior (respecting the different traffic rules), $r_{\text{unsafe}} = 10$ for unsafe behavior, and $r_{\text{critical}} = 50$ for severe violations.

This summarizes the reward structures for the three driving
requests (\textit{fast}, \textit{comfort}, \textit{stop}).

\section{LLM Modules Design}
\label{sec:Section4}

This section describes the design and roles of the LLM-based modules used in the proposed framework, as well as the prompts that are passed to each module.

\subsection{Command Interpretation Module}

This module, named LLM-1, converts a passenger request, whether direct or indirect, into a single unambiguous driving instruction among three styles: \textit{drive faster}, \textit{drive more carefully}, or \textit{stop}. To avoid misunderstandings, LLM-1 always reformulates the request and asks the passenger for confirmation before any action is applied. The following prompt was used for this task:
\begin{quote}
\small
\texttt{You are an autonomous taxi's AI assistant. Reformulate the passenger's request into a direct instruction about how the vehicle should drive. The instruction must correspond to one of the following driving styles: drive faster, drive more carefully, or stop. At the end, confirm the reformulation using the following format: ``You want me to [action], is this correct?'' Do not provide multiple options or add any extra text. Only output the direct confirmation question.}
\end{quote}
For each passenger command, LLM-1 outputs only the confirmation question indicating which driving request to execute. The command is then mapped to a numeric class used by downstream controllers:

drive faster $\rightarrow$ request \textbf{0}; \quad
drive more carefully $\rightarrow$ request \textbf{1}; \quad
stop $\rightarrow$ request \textbf{2}.

This design produces a user-verified intention and a compact, machine-readable output (cmd $\in \{0,1,2\}$) that can be directly used by the RL agent.

\subsection{Decision Module}

This second module, named LLM-2, leverages the LLM’s reasoning capability to determine whether the vehicle should proactively explain its current behavior to the passenger. The decision is based on three inputs provided at each decision step:
\begin{itemize}
    \item \textbf{Passenger Request}: the latest confirmed intent,
    \item \textbf{Vehicle Action}: a short textual description of the maneuver selected by the RL agent from the action space defined in Eq. (\ref{eq:action_space}),
    \item \textbf{Environment State}: a human-readable summary generated by a dedicated function, including ego speed, speed limit, area type (urban, residential), distance to the leading vehicle, distance and color of the next traffic light, and distance to the next pedestrian crossing. Objects further than 50\,m are ignored to keep the LLM focused on near-term safety.
\end{itemize}
The model operates at a low temperature and is constrained to output only one of two predefined tags. The output is then trimmed and encoded as \{[EXPLAIN: 1], [NO EXPLANATION NEEDED: 0]\} for downstream processing. The following is the prompt that is passed to LLM-2:
\begin{quote}
\small
\begin{verbatim}
You are an AI reasoning module for an autonomous taxi. Your task is to output exactly ONE of:
- [EXPLAIN]
- [NO EXPLANATION NEEDED]

Principles to decide:

1) Return [EXPLAIN] only if the vehicle’s present action clearly conflicts
or delays the user’s request for a clear safety reason (e.g., decelerating
when asked for driving faster, or harsh acceleration or deceleration when
asked for a comfortable ride).

2) Return [NO EXPLANATION NEEDED] for behavior that is ordinary, supportive,
or neutral with respect to the request (e.g., maintaining speed when user
asked for comfort, smooth acceleration/braking).

3) Multiple reasons: If several safety reasons exist, internally consider
the most safety-critical one.
Output only the tag. Do not include any justification or text besides the
tag.
\end{verbatim}
\end{quote}

\subsection{Explanation Generator}

When LLM-2 returns \textit{[EXPLAIN]}, the explanation generator module, named LLM-3, produces a short, user-friendly justification of the vehicle’s behavior. It relies on the same inputs as the decision module and generates a one- to two-sentence explanation in a friendly and reassuring tone. LLM-3 is invoked only when an explanation is required. It operates at a low temperature (0.2) and selects the most safety-relevant nearby factor from the environment summary, such as a red traffic light, a pedestrian crossing, or a short headway, and explains the behavior concisely. For this, we used the following prompt:

\begin{quote}
\small
\begin{verbatim}
You are an AI assistant inside an autonomous taxi. Your job is to explain
to the passenger why the vehicle’s action seems to contradict their request.
When that happens, briefly explain the most safety-relevant nearby reason
for the vehicle’s behavior, using a friendly and reassuring tone. Keep
it concise.
\end{verbatim}
\end{quote}

\section{Evaluation and Results}
\label{sec:Section5}
This section evaluates the proposed framework across three dimensions: 
\begin{enumerate}
    \item the training performance of the DRL agents,
    \item their ability to execute safe and efficient driving under different passenger commands, and
    \item the effectiveness of the LLM modules in interpreting requests, detecting conflicts, and generating clear explanations.
\end{enumerate}

The agent was trained using D3QN. The neural network employs two fully connected hidden layers with 128 units each. It is trained with the Adam optimizer (learning rate $10^{-3}$) and mean squared error loss. Training was conducted for 11,000 episodes using a replay buffer with a capacity of 5M transitions, a mini-batch size of 64 samples and a discount factor $\gamma = 0.99$. In even-numbered episodes, the primary driving command (\textit{fast} or \textit{comfort}) was periodically interleaved with the stop command, while in odd-numbered episodes it was applied throughout the entire episode. This ensures regular exposure to emergency stopping scenarios under varied traffic conditions.

\subsection{Training Performance Analysis}

\begin{figure}[thbp]
    \centering

    \begin{subfigure}[b]{0.49\columnwidth}
        \centering
        \includegraphics[width=\textwidth, height=0.18\textheight]{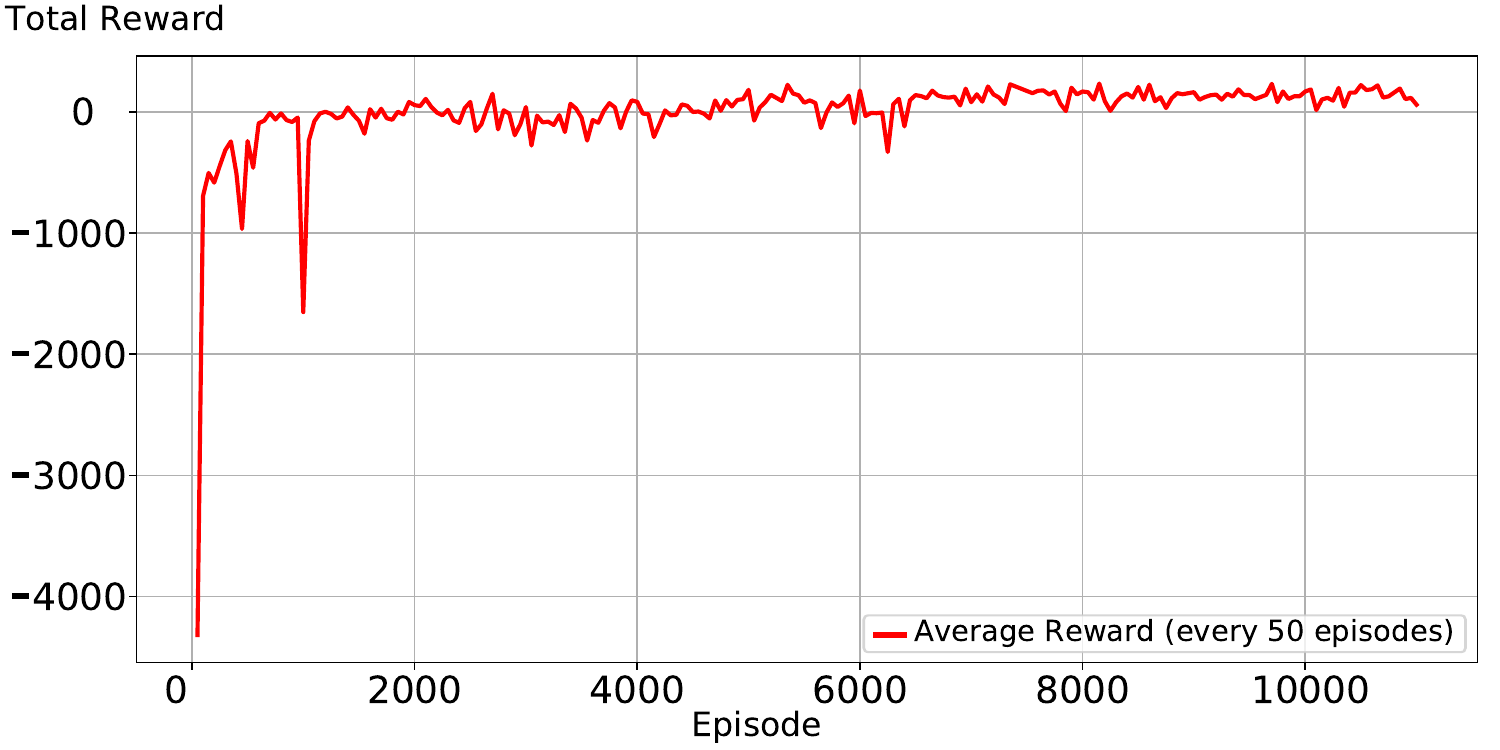}
        \caption{Fast driving model}
        \label{fig:train_fast}
    \end{subfigure}
    \hfill
    \begin{subfigure}[b]{0.49\columnwidth}
        \centering
        \includegraphics[width=\textwidth, height=0.18\textheight]{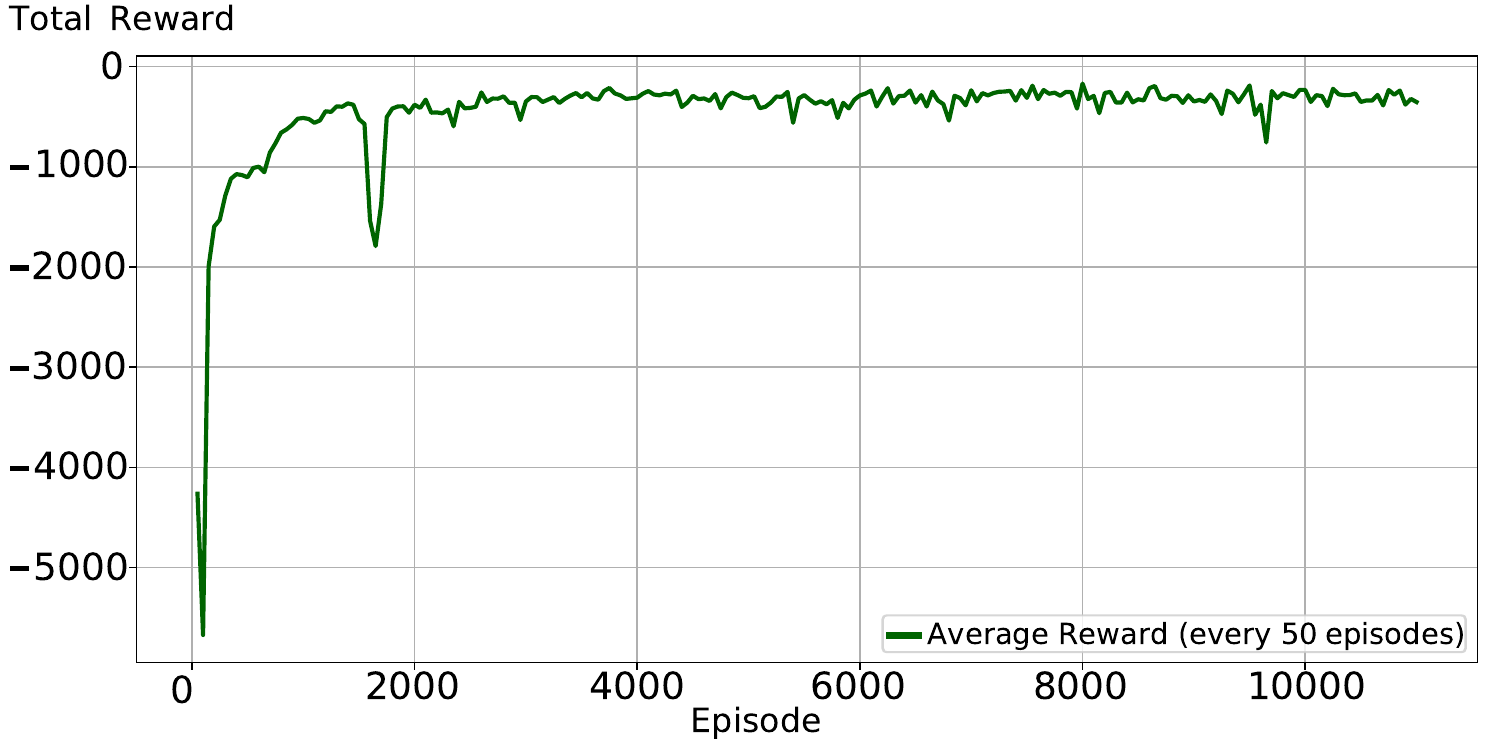}
        \caption{Comfort driving model}
        \label{fig:train_comfort}
    \end{subfigure}

    \caption{Training rewards with 50-episode moving average for the fast and comfort driving policies.}
    \label{fig:train_combined}

\end{figure}

Figures~\ref{fig:train_fast} and~\ref{fig:train_comfort} show the training curves of the fast and comfort-oriented agents, reporting total episodic reward averaged over 50 episodes, where rewards start low due to exploration, then quickly stabilize at positive values, indicating stable learning with the D3QN architecture and proposed reward shaping.

\subsection{Decision-Making Performance and Behavioral Analysis}
\label{sec:decision_behavior}

This subsection analyzes the driving behavior and safety
performance of the proposed framework for the fast driving agent, the comfort-oriented driving agent, and the merged multi-command configuration.

Evaluation is based on speed-time curves and aggregate safety metrics computed over multiple episodes. Specifically, both trained policied were evaluated using five different seeds, with each seed tested across five traffic levels. The traffic flow levels are explicitly set to $0.5$, $0.75$, $1.0$, $1.25$, and $1.5$. The baseline level $1.0$ corresponds to inserting one vehicle every 4 seconds (equivalent to 900 vehicles/hour) in the main road used by the ego-vehicle. Therefore, traffic levels $0.5$, $0.75$, $1.25$, and $1.5$ correspond to approximately $450$, $675$, $1125$, and $1350$ vehicles/hour, respectively. This results in a total of 25 evaluation cases (5 seeds × 5 traffic levels).

\subsubsection{Fast Driving Agent}
\begin{figure}[!htbp]
    \centering

    \begin{subfigure}[b]{\columnwidth}
        \centering
        \includegraphics[width=0.75\textwidth]{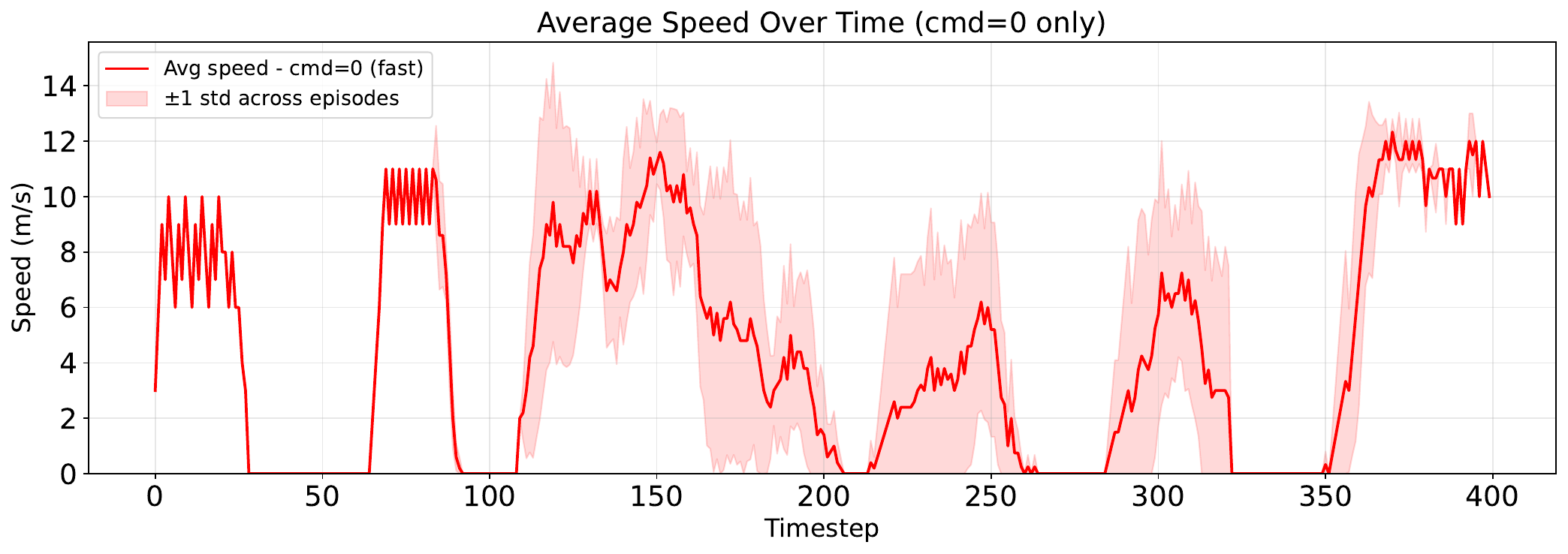}
        \caption{Fast driving only}
        \label{fig:fast}
    \end{subfigure}

    \begin{subfigure}[b]{\columnwidth}
        \centering
        \includegraphics[width=0.75\textwidth]{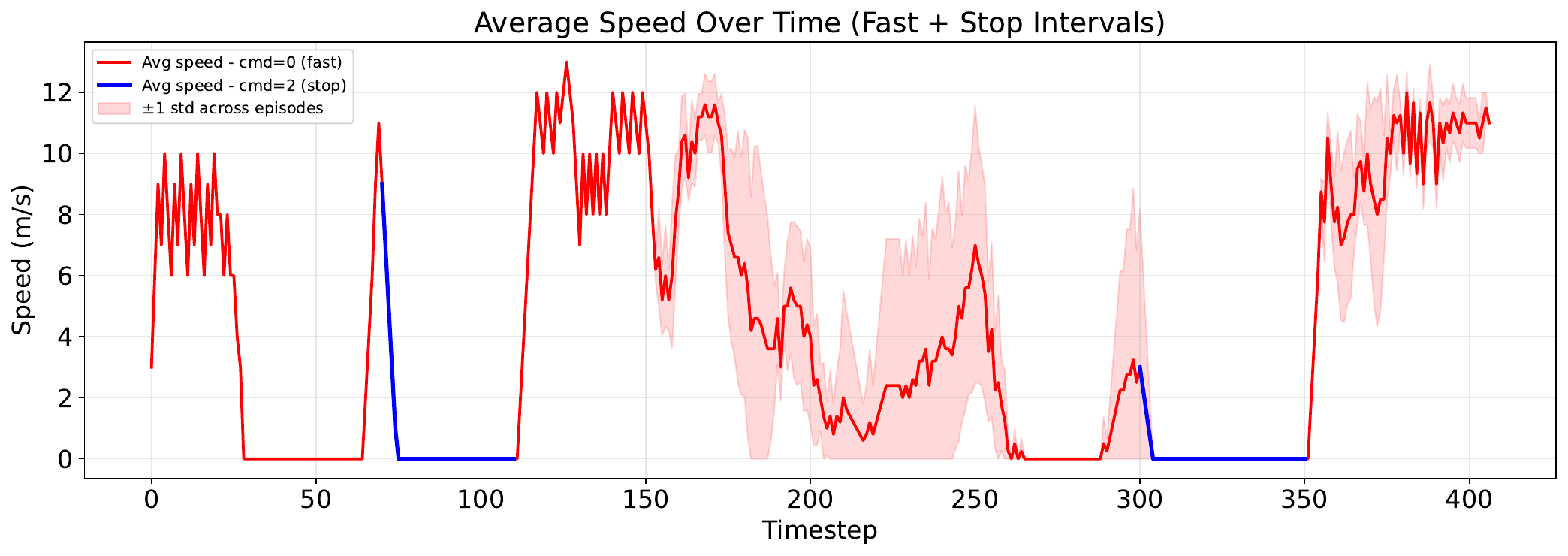}
        \caption{Fast driving with stop commands}
        \label{fig:fast_stop}
    \end{subfigure}

    \caption{Ego-vehicle speed profiles averaged across 25 episodes (5 seeds × 5 traffic levels) under a fast driving policy, with (a) and without (b) stop command injection at fixed intervals. The shaded region represents ±1 standard deviation across episodes.}
    \label{fig:speed_combined}

\end{figure}

Figure ~\ref{fig:fast} presents the average ego-vehicle speed over all 25 evaluation episodes executed under the fast driving command only, with the shaded region representing $\pm1$ standard deviation across runs. The speed profile shows that the trained policy generally accelerates at the beginning of the episode, reaches higher speeds during free-flow segments, and then experiences repeated speed reductions. This is due to interactions with leading vehicles and traffic constraints, which causes frequent braking and re-acceleration, resulting in visible oscillations in the speed profile. This behavior confirms that the fast driving policy prioritizes efficiency and progress over smoothness, while still responding appropriately to safety constraints such as red lights, pedestrian crossings, and car following.

When stop commands are injected during fast driving
(Fig.~\ref{fig:fast_stop}), the agent consistently executes a full stop within the commanded intervals and resumes fast driving once the stop command is released. This demonstrates that explicit safety requests can temporarily override efficiency-oriented behavior.

Quantitative safety evaluation of this model was also conducted over the 25 evaluation episodes. As shown in Table~\ref{tab:fast_model_evaluation}, the agent maintains an overall average time headway TH of 2.998~s and an average TTC of 5.603~s, while achieving an average speed of 4.508~m/s. Across all traffic levels, no collisions involving the autonomous vehicle were observed. These results indicate that the fast-driving policy is able to maintain relatively high speeds while preserving safe longitudinal interactions with surrounding vehicles under the evaluated traffic conditions.

\begin{table}[htbp]
\centering
\caption{Evaluation results of the fast-driving model under different traffic levels.}
\label{tab:fast_model_evaluation}
\resizebox{\columnwidth}{!}{
\begin{tabular}{c c c c c c}
\hline
\textbf{Traffic level} & 
\textbf{SUMO steps} & 
\textbf{Mean speed (m/s)} & 
\textbf{Avg. TH (s)} & 
\textbf{Avg. TTC (s)} & 
\textbf{AV collisions} \\
\hline
0.50 & 253 & 6.087 & 3.093 & 6.669 & 0 \\
0.75 & 323 & 4.767 & 3.387 & 6.106 & 0 \\
1.00 & 386 & 3.982 & 3.096 & 4.302 & 0 \\
1.25 & 401 & 3.825 & 2.861 & 4.755 & 0 \\
1.50 & 397 & 3.879 & 2.553 & 6.184 & 0 \\
\hline
\multicolumn{2}{c}{\textbf{Overall average}} &
4.508 & 2.998 & 5.603 & 0 / 25 \\
\hline
\end{tabular}
}
\end{table}

\subsubsection{Comfort-Oriented Driving Agent}

\begin{figure}[thbp]
    \centering

    \begin{subfigure}[b]{0.8\columnwidth}
        \centering
        \includegraphics[width=\textwidth]{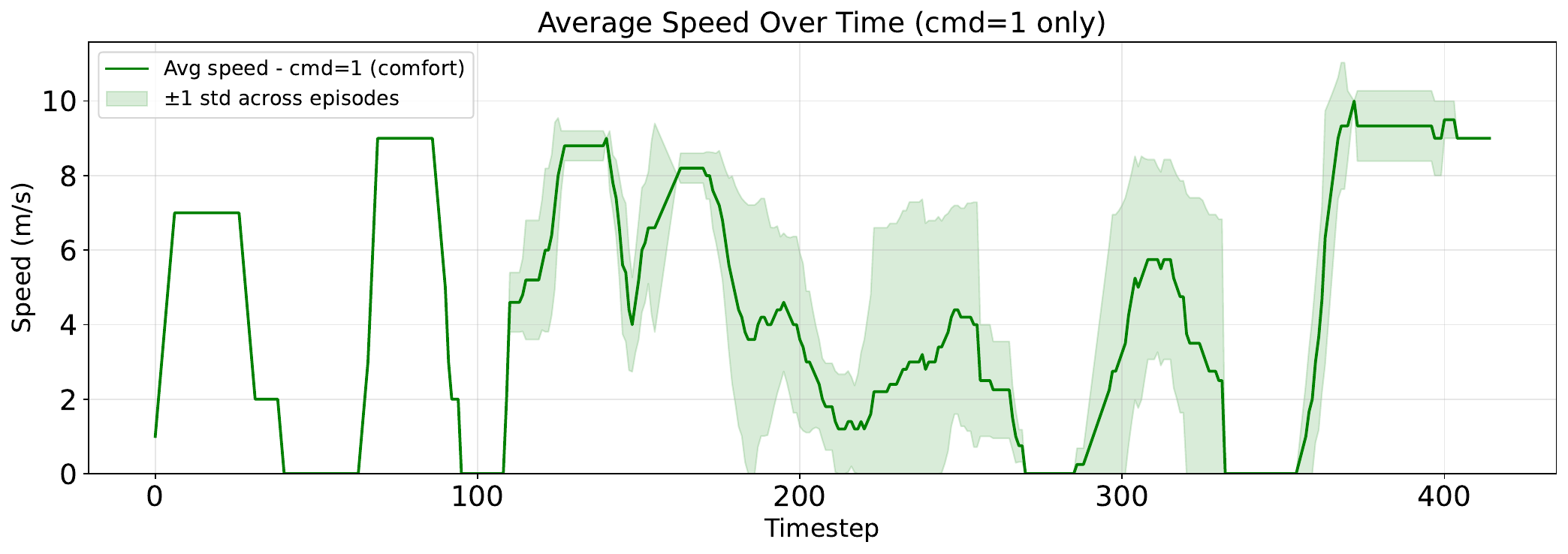}
        \caption{Comfort driving only}
        \label{fig:comfort}
    \end{subfigure}

    \begin{subfigure}[b]{0.80\columnwidth}
        \centering
        \includegraphics[width=\textwidth]{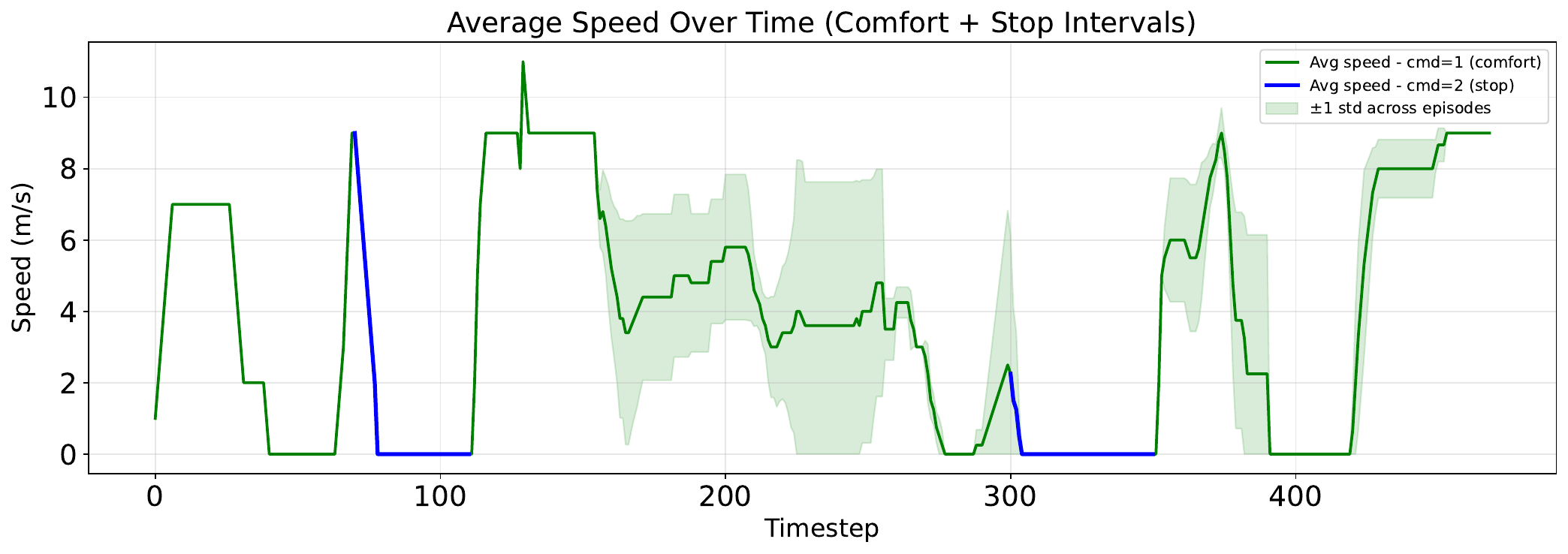}
        \caption{Comfort driving with stop commands}
        \label{fig:comfort_stop}
    \end{subfigure}

    \caption{Ego-vehicle speed profiles averaged across 25 episodes (5 seeds × 5 traffic levels) under a comfort driving policy, with (a) and without (b) stop command injection at fixed intervals. The shaded region represents ±1 standard deviation across episodes.}
    \label{fig:comfort_combined}

\end{figure}

Figure~\ref{fig:comfort} illustrates the behavior of the comfort-oriented agent under a comfort command only, across all 25 episodes. In contrast to the fast driving agent (Fig. \ref{fig:speed_combined}), the comfort driving policy produces smoother speed profiles with fewer oscillations and lower cruising speeds. Acceleration and deceleration phases are more gradual, reflecting the explicit penalization of jerk and aggressive maneuvers in the reward function.

With scheduled stop commands injected into comfort driving
(Fig.~\ref{fig:comfort_stop}), the agent reliably performs complete stops within the commanded intervals and resumes comfort-oriented driving afterward.

Aggregate results in table \ref{tab:comfort_model_evaluation}, over the 25 episodes, show that the comfort-driving model achieves an overall average speed of 4.396~m/s, which is slightly lower than the fast-driving model average of 4.508~m/s. However, the comfort policy maintains a substantially larger average time headway TH of 6.373~s compared to 2.998~s for the fast model, indicating more conservative longitudinal behavior. The average TTC remains comparable, with 5.460~s for the comfort model and 5.603~s for the fast model. In addition, the comfort model achieves a low average jerk of 0.205~m/s$^3$, reflecting smoother driving behavior. No AV collisions were observed across all 25 episodes, similarly to the fast-driving model. Overall, these results show that the comfort policy sacrifices a small amount of speed in favor of larger headway distances and smoother motion.

\begin{table}[htbp]
\centering
\caption{Evaluation results of the comfort-driving model under different traffic levels.}
\label{tab:comfort_model_evaluation}
\resizebox{\columnwidth}{!}{
\begin{tabular}{c c c c c c c}
\hline
\textbf{Traffic level} & 
\textbf{SUMO steps} & 
\textbf{Mean speed (m/s)} & 
\textbf{Avg. TH (s)} & 
\textbf{Avg. TTC (s)} & 
\textbf{Avg. jerk (m/s$^3$)} & 
\textbf{AV collisions} \\
\hline
0.50 & 257 & 5.996 & 4.377 & 3.745 & 0.219 & 0 \\
0.75 & 333 & 4.617 & 5.579 & 4.654 & 0.169 & 0 \\
1.00 & 398 & 3.869 & 6.659 & 5.578 & 0.232 & 0 \\
1.25 & 416 & 3.699 & 7.986 & 6.916 & 0.198 & 0 \\
1.50 & 405 & 3.800 & 7.264 & 6.405 & 0.208 & 0 \\
\hline
\multicolumn{2}{c}{\textbf{Overall average}} &
4.396 & 6.373 & 5.460 & 0.205 & 0 / 25 \\
\hline
\end{tabular}
}
\end{table}

\subsubsection{Merged Multi-Command Evaluation}

\begin{figure}[ht]
    \centering
    \includegraphics[width=1\columnwidth]{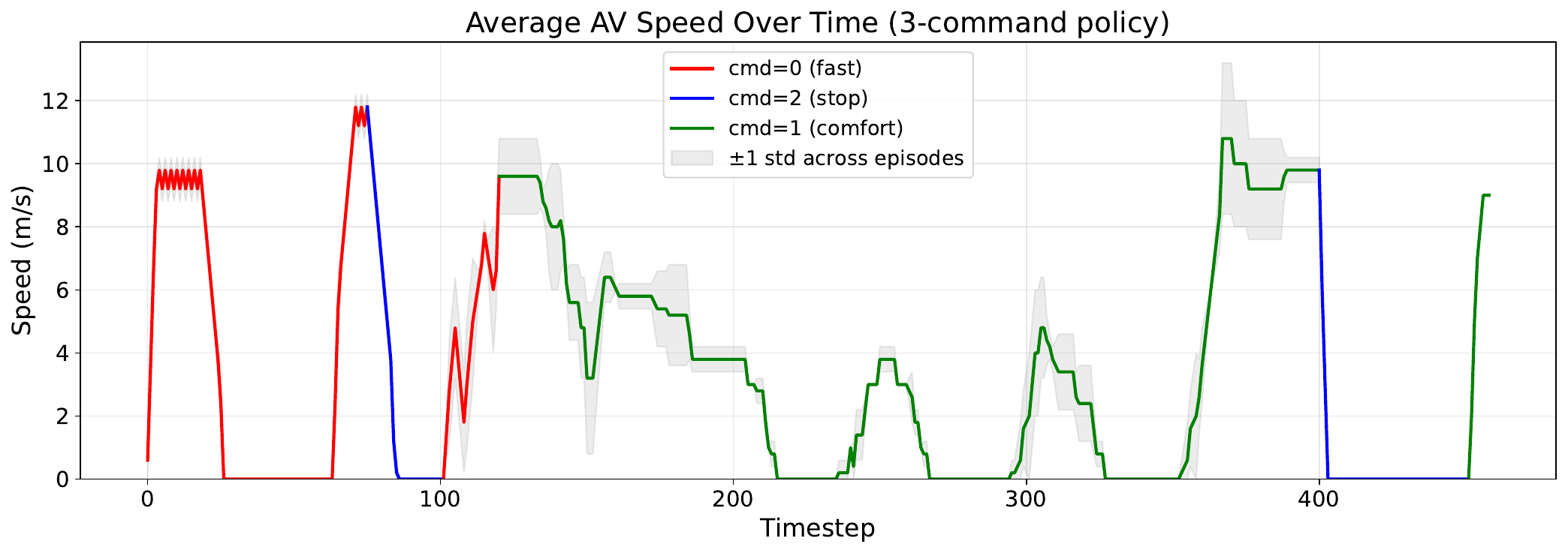}
    \caption{Merged-model single-episode evaluation.}
    \label{fig:merged}
\end{figure}

To evaluate dynamic switching between driving modes, the fast and comfort driving agents were integrated into a unified framework and tested across the 25 episodes, involving sequential fast, stop, and comfort commands. Figure~\ref{fig:merged} shows that the agent adapts its behavior correctly to each command. Under the fast command, the vehicle accelerates aggressively; when a stop command is issued, it reliably comes to a complete halt; and when switched to the comfort command, it transitions to smoother acceleration and lower cruising speeds. These transitions occur without instability or safety violations.

Overall, this experiment demonstrates that the proposed framework supports seamless runtime switching between driving styles while preserving safety, rule compliance, and responsiveness to explicit passenger commands.

\subsection{Explainability Performance Evaluation}
\label{sec:xai_eval}

This subsection evaluates the performance of the LLM-based explainability modules, including passenger command interpretation (LLM-1), explanation triggering (LLM-2), and explanation generation (LLM-3).

\subsubsection{Passenger Command Interpretation (LLM-1)}

The evaluation of this first module was conducted on a curated dataset of 33 passenger requests, including both direct and indirect formulations. Outputs were required to follow the exact confirmation template and contain a single valid action. Results show that it correctly classified 32 out of 33 requests, achieving an accuracy of 0.969, a macro-F1 score of 0.970, and a Cohen’s $\kappa$ of 0.954. Template compliance was 1.0, indicating fully parseable outputs. Class-wise results (Table~\ref{tab:llm1_class}) show perfect separation for \textit{FAST}, while the only error corresponds to a \textit{STOP} request misclassified as \textit{COMFORT}. Although rare, this error is safety-relevant and motivates prioritizing STOP recall in future work.

\begin{table}[h]
\centering
\caption{LLM-1 per-class classification performance}
\label{tab:llm1_class}
\begin{tabular}{lcccc}
\hline
\textbf{Class} & \textbf{Precision} & \textbf{Recall} & \textbf{F1} & \textbf{Support} \\
\hline
FAST & 1.000 & 1.000 & 1.000 & 10 \\
COMFORT & 0.923 & 1.000 & 0.960 & 12 \\
STOP & 1.000 & 0.909 & 0.952 & 11 \\
\hline
\end{tabular}
\end{table}

\subsubsection{Decision and Explanation Modules (LLM-2 \& LLM-3)}

The decision and explanation modules were evaluated qualitatively during closed-loop driving episodes. Explanations were correctly triggered when vehicle actions deviated from passenger requests due to safety constraints, such as red traffic lights, pedestrian crossings, or short headways.

In fast-driving scenarios, explanations reliably justified safety-driven decelerations or limited acceleration. For example, when the vehicle slowed down despite an active fast-driving request due to an upcoming traffic light, the system generated the following explanation:
\begin{quote}
\small
\texttt{I'm slowing down slightly because we're approaching a traffic light just ahead. This helps ensure a smooth and safe stop if needed. Your safety is my top priority!}
\end{quote}

Under comfort-oriented requests, explanations were most effective during non-trivial maneuvers, such as stronger acceleration to merge smoothly with traffic. In one representative case, the explanation module produced:
\begin{quote}
\small
\texttt{I'm accelerating a bit to smoothly merge with traffic ahead, ensuring a safer and more comfortable ride overall.}
\end{quote}

While these examples demonstrate that the explainability pipeline provides clear, context-aware, and passenger-oriented justifications for meaningful control decisions, explanations triggered for minor adjustments (e.g., small speed changes or steady-speed maintenance) provided limited added value. This suggests that further reducing explanation redundancy for low-impact actions could improve overall user experience.

\section{Discussion and Conclusion}
\label{sec:Section6}

This paper presents a proof-of-concept hybrid autonomous driving framework that combines multiple DRL driving policies with LLM-based modules for passenger command interpretation, conflict detection, and natural language explanation generation. 

Experimental results in a realistic SUMO urban environment show that the DRL agents achieve stable learning, respect traffic rules, and adapt their behavior to \textit{fast} driving, \textit{comfort-oriented} driving, and stop commands, including dynamic command switching within a single episode, while quantitative safety metrics confirm collision-free operation. Moreover, the explainability pipeline generates context-aware, safety-oriented justifications when vehicle behavior deviates from passenger expectations. 

Despite these promising results, several limitations remain, as all experiments were conducted in simulation using SUMO, and real-world deployment would introduce additional challenges related to sensor noise, perception uncertainty, and real-time latency. The evaluation of the explanation modules (LLM-2 and LLM-3) was qualitative due to the lack of standardized automatic metrics for real-time passenger-oriented explanations, and the command interpretation evaluation relied on a relatively small dataset, requiring broader testing to fully assess robustness to linguistic diversity. Furthermore, while low-temperature decoding reduces hallucinations, LLM outputs may still exhibit occasional inconsistencies, highlighting the need for stronger grounding mechanisms and conservative fallback strategies in safety-critical scenarios. Future work will extend this work to lateral control tasks, including lane changing and more complex multi-lane traffic scenarios.

In conclusion, this work demonstrates the feasibility of integrating DRL-based driving control with LLM-driven passenger interaction and explainability within a unified framework. The results indicate that such integration supports adaptive driving behavior, safe rule compliance, and context-aware communication with passengers. While limited to a simulation environment, this study provides a foundation for future extensions toward more complex scenarios, including lateral control and real-world deployment, and underscores the potential of language-driven reasoning to enhance transparency and trust in autonomous driving systems.

\bibliographystyle{unsrtnat}
\bibliography{references}  






\end{document}